\documentclass[10pt,twocolumn,letterpaper]{article}

\usepackage[pagenumbers]{wacv} 

\usepackage{graphicx}
\usepackage{amsmath}
\usepackage{amssymb}
\usepackage{booktabs}

\usepackage[pagebackref,breaklinks,colorlinks]{hyperref}

\usepackage[capitalize]{cleveref}
\crefname{section}{Sec.}{Secs.}
\Crefname{section}{Section}{Sections}
\Crefname{table}{Table}{Tables}
\crefname{table}{Tab.}{Tabs.}

\def\wacvPaperID{1313} 
\def\confName{WACV}
\def\confYear{2024}

\begin{document}

\title{SENetV2: Aggregated dense layer for channelwise and global representations}

\author{Mahendran N\\
{\tt\small mahendranNNM@gmail.com}
}
\maketitle

\begin{abstract}
   Convolutional Neural Networks (CNNs) have revolutionized image classification by extracting spatial features and enabling state-of-the-art accuracy in vision-based tasks. The squeeze and excitation network proposed module gathers channelwise representations of the input. Multilayer perceptrons (MLP) learn global representation from the data and in most image classification models used to learn extracted features of the image. In this paper, we introduce a novel aggregated multilayer perceptron, a multi-branch dense layer, within the Squeeze excitation residual module designed to surpass the performance of existing architectures. Our approach leverages a combination of squeeze excitation network module with dense layers. This fusion enhances the network's ability to capture channel-wise patterns and have global knowledge, leading to a better feature representation. This proposed model has a negligible increase in parameters when compared to SENet. We conduct extensive experiments on benchmark datasets to validate the model and compare them with established architectures. Experimental results demonstrate a remarkable increase in the classification accuracy of the proposed model.

\end{abstract}

\section{Introduction}
\label{sec:intro}
Deep learning progress has led to remarkable advancements in various domains, particularly in vision based tasks. Convolutional Neural Networks (CNNs) have played a pivotal role in image classification, object detection, and feature extraction. The complexity of datasets and challenges continues to grow, there is a persistent pursuit to further enhance the accuracy of CNNs. Diverse architectural innovations have arisen through a step-by-step process, identifying challenges and subsequently providing solutions \cite{inception,resnet1000}. Following the success of AlexNet \cite{alexnet}, an array of architectures have surfaced, notably including VGG \cite{vgg2014} and the Residual network \cite{resnet}. Convolutions-based architectural approaches have been explored extensively to enhance the network's representational power \cite{senet}. To enhance performance, researchers pursuit has led to the exploration of novel architectural modifications, specialized layers, and fusion techniques to extract the most informative features from input dataset. Each step in this journey has brought about new paradigms, novel design principles, and groundbreaking techniques that have not only pushed the boundaries of what neural networks can achieve but have also sparked new avenues of research.

Convolutional neural networks (CNNs) excel in learning spatial correlations within localized receptive fields. CNNs learn new features, thereby contributing to model performance. On the other hand, Fully Connected (FC) layers specialize in learning global representations by virtue of their connections with all nodes within the layer. FC layers often find their place in architectures toward the final stages, aiding in the classification process. The Squeeze and Excitation Network (SENet), one of the notable advancements, explicitly models interdependencies among channels \cite{senet}. This deliberate modeling significantly enhances the network's representational potency via feature recalibration. SENet \cite{senet} introduces internal channel-level adjustments that impact the model's performance. This module effectively transmits crucial information from globally learned insights to the subsequent layer, thereby enhancing the performance. The Squeeze and Excitation network uses the concept of channelwise representations within the model, further expanding its capabilities.

Christian et al. \cite{inception} introduced the Inception module, a novel architectural concept which incorporates multibranch convolutions with varying filter sizes, which are subsequently concatenated at the block's conclusion. This inventive strategy ignites a fresh architectural paradigm, presenting a methodology to attain enhanced performance with reduced theoretical complexity. CNNs are typically incorporated within the inception module to facilitate spatial representation learning. Similar strategies have been proposed, such as ResNeXt, which integrates multiple branches with identical topology within a module. These amalgamated modules are also characterized by a new dimension known as cardinality. The incorporation of aggregated networks enhances the model's ability to learn spatial correlations without necessitating increased depth, optimizing computational resources. Subsequent researchers have embraced the inception-based approach, resulting in the creation of diverse network architectures \cite{resnext,xception,wrns,mobilenet}.

\begin{figure*}[hbt!]
    \centering
    \includegraphics[width=\textwidth,height=10cm]{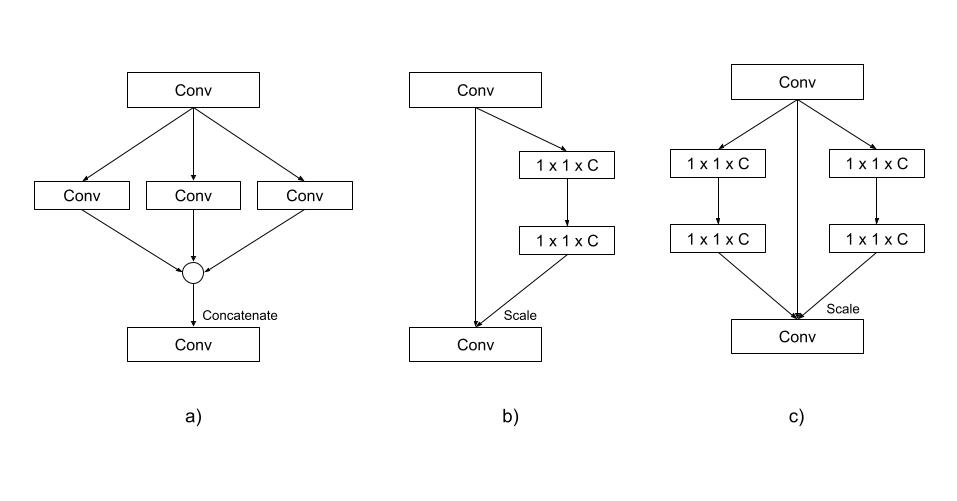}
    \caption{The comparison among the ResNeXt, SENet, and the proposed SENetV2 modules is presented as follows.a) The ResNext aggregated module with multi branch CNNs. b) The SE module involves reducing the input and then restoring its original shape. In the case of the squeeze operation, the reduced input is not sent back to regain the original shape. c) The proposed SENetV2 module is a combination of a) and b) consisting of multi branch FC layers.}
    \label{fig:mainimages}
\end{figure*}

SENet enhances channelwise representations via squeeze and excitation operations, thus bolstering their expressive capabilities. Channelwise features undergo recalibration through the application of the squeezed input. The excitation component within the proposed module involves a fully connected layer that captures global representations. This specialized module ensures that solely the most pivotal features are transmitted to the ensuing layer. When this distinct module design is iteratively applied within networks like ResNet, specifically within residual modules, its impact magnifies, essentially acting as a filtering mechanism for the network, as seen in SE ResNet.

\begin{figure}
    \centering
    \includegraphics[width=0.9\linewidth,height=10cm]{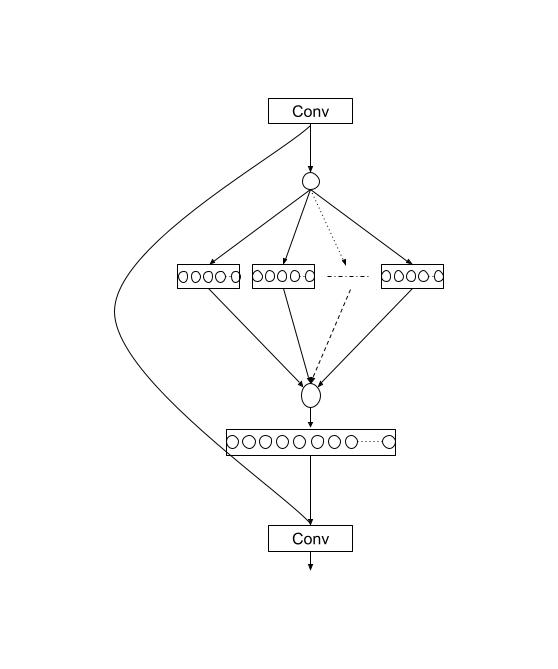}
    \caption{The figure illustrates the internal functioning of the proposed SaE module of SENet V2. The squeezed output is then fed into multi branch FC layers, followed by the excitation process. The split input is passed to the end to restoring its original shape.}
    \label{fig:complete}
\end{figure}

We propose an upgraded SE module, Squeeze aggregated excitation (SaE), to be incorporated in SENet. The comparison among the Aggregated Residual Module, the Squeeze and Excitation Module, and the proposed Squeeze Aggregated Excitation (SaE) Module is depicted in Figure \ref{fig:mainimages}. The figure illustrates that both the Squeeze Excitation Module and the SaE Module selectively transmit crucial features. However, the SaE Module optimizes this stage by increasing the cardinality between layers. Drawing inspiration from the inception module, we introduce an multibranch FC layer of the same size, akin to the approach of ResNeXt. This decision is guided by its amplified impact as previously discussed \cite{resnext}.  The architecture of the proposed module is visualized in Figure \ref{fig:complete}. The theoretical complexity of the proposed module surpasses that of the existing Squeeze and Excitation Module. We adopt a reduction size of 32 and a cardinality of 4, deliberately keeping the cardinality modest to prevent unnecessary complexity increase. The results emanating from the aggregated FC layers are concatenated within the excitation phase, reinstating the desired output shape for subsequent layers.

The main contributions of this paper is summarized as follows:
\begin{itemize}
    \item We acknowledge the significant influence of aggregated modules and the robust representational capabilities inherited from Inception architectures and the squeeze and excitation module, respectively. The integration of aggregated modules contributes to a reduction in the theoretical complexity of the network.
    \item Building on the concept that fully connected layers grasp insights from the network's global representation, we propose an intriguing notion: to utilize fully connected layers as aggregated modules, a concept substantiated in \cite{resnext}.
    \item Our approach entails the introduction of a multi-branch configuration for fully connected layers, wherein the squeezed layer imparts critical features prior to excitation, thus restoring its original form. This aggregated module not only boasts reduced complexity but also outperforms conventional aggregated modules by selectively incorporating essential features while disregarding others.
\end{itemize}

\section{Related Work}
The preceding research pertaining to the proposed model is delineated within this section. Commencing with Yann LeCun's pioneering work on ConvNets, convolutional neural networks (CNNs) have achieved remarkable success in diverse vision applications. Subsequent advancements emerged as computational capabilities improved, providing researchers with enhanced tools to contribute significantly to the field of network engineering. Noteworthy architectures leveraging CNNs for improved performance include AlexNet \cite{alexnet}, ZFNet \cite{zfnet}, VGG \cite{vgg2014}, and ResNet \cite{resnet}. 

In 2015, the inception architecture achieved competitive performance levels while maintaining reduced computational complexity. The methodology employed multi-branch convolutions, tailoring convolutions within each branch. By harnessing this approach, the architecture maximized the utilization of image modules, incorporating numerous convolutional filters within its multi-branched structure. Inception's innovative approach secured victory in the ILSRVC in 2014 and notably reduced the parameter count from 60 million in the previous best-performing AlexNet to a mere 4 million.

This concept was subsequently adopted in aggregated ResNet modules. Notable architectures that have emerged in the wake of Inception, either directly or indirectly drawing traits from it, include ResNeXt, Xception, and Mobilenet \cite{resnext,mobilenet,xception}. The inception module underscored the advantages of delving into deeper networks. The Xception architecture delved further by fragmenting the convolution operations within the inception module, resulting in considerably swifter convolution operations \cite{xception}. Building on Xception's foundation, Mobilenet introduced the utilization of depthwise separable convolutions across all layers, thereby significantly reducing computations and leading to a more compact model size \cite{mobilenet}.

Grouped convolutions are instrumental in the distribution of models across multiple GPUs \cite{alexnet}. However, there is limited evidence to suggest that this convolutional approach significantly enhances accuracy, as far as our current understanding extends. Channel-wise convolution constitutes a variant derived from grouped convolutions, where the number of groups aligns with the number of channels in the input data.

Residual networks (ResNets) have introduced a revolutionary approach to address the challenge of vanishing gradients. In architectures with significant depth, model performance deteriorates due to the difficulty of conveying learned representations effectively across the numerous layers. The innovative solution involves the implementation of residual modules, integrating shortcut connections. These connections facilitate the continuous propagation of previously acquired representations at regular intervals, mitigating the degradation in performance associated with deep networks. Resnets can be expanded even up to 1000 layers \cite{resnet1000}. An instrumental component in stabilizing the learning process within ResNets is batch normalization \cite{batchnorm}. This technique aids normalize inputs on a batch-wise basis and significantly contributes to the network's ability to converge efficiently during training.

ResNeXt \cite{resnext} have Aggregated Residual Module by introducing a new dimension called "cardinality". The adoption of cardinality reduces theoretical complexity while achieving superior performance compared to traditional stacked architectures, as posited by Saining \cite{resnext}. The Wide Residual Network (WRN) \cite{wrns} is another variant of the residual network that focuses on enhancing the width of convolutional layers. This expansion augments the module's learning capacity beyond the confines of traditional ResNet. Researchers also proposed ResNeSt \cite{resnest}, which leverages split attention networks. In this framework, the module constructs a multipath network by partitioning layers using a 1x1 convolution, followed by a 3x3 convolution. These paths are then merged using a split attention block, combining channel-wise representations effectively.

SENet \cite{senet}, on the other hand, introduces a ResNet-based squeeze and excitation module. This novel module design significantly contributes to increased model performance by incorporating channel-wise representations. Highway networks are characterized by a gated mechanism regulating shortcut connections \cite{highway}. There exists a significant relationship between the utilization of various filters within architectural designs, as this approach enhances the network's representational capacity by capturing finer patterns. Noteworthy strategies, such as Inception \cite{inception} and Pyramid Network \cite{pyramid}, incorporate diverse topologies to yield improved performance outcomes.

In architectures like Xception \cite{xception} and MobileNets \cite{mobilenet}, the concept of depthwise separable convolution is employed. This convolutional technique comprises two distinct stages: depthwise convolution and pointwise convolution. In depthwise convolution, a singular filter is applied to each input channel. Subsequently, the pointwise convolution involves the application of a 1x1 kernel filter to each value within the representation. An additional component, known as channel-wise convolution, is embedded within the separable convolution process. In this phase, the convolutions related to individual channels are executed before the pointwise convolutions take place, contributing to the network's efficient feature extraction.

These architectures indeed exhibit reduced parameter counts when compared to similar-performing architectures or non-aggregated networks. This work aligns with the concept of the squeeze operation, strategically streamlining the network's structure and its complexity. Various other architectures, such as Structured Transform Networks \cite{structured}, Deep Fried ConvNets \cite{deepfried}, and ShuffleNet \cite{shufflenet}, also consider computational efficiency as a pivotal aspect. These architectures prioritize reduced computation, compact networks, and some are optimized for mobile applications. The proposed method demonstrates a marginal increase in parameters, yet significantly enhances the model's accuracy by improving its ability to distinguish between different classes.

Another pertinent area related to the proposed approach is that of model compression methods. Several alternative strategies for constructing smaller networks also exist, often achieved through compression methods such as quantization of pretrained networks. The attention mechanism within CNNs is designed to emphasize the most relevant portions of an image while disregarding irrelevant portions. This approach aids in effectively comprehending complex scenes within an image. Typically, attention mechanisms employ softmax or sigmoid functions to serve as gating mechanisms. Within the context of the SEnet, the attention mechanism operates within the SE block, focusing on channel-wise relationships. This akin approach is adopted within the proposed SaE module in SENetV2 network, incorporating a lightweight gating mechanism to highlight channel-wise relationships.

Ensembles encompass multiple replicas of the same model operating in parallel to address the same problem, with the final result being determined through a collective consideration of all outcomes. Ensemble models often surpass single models in terms of performance.

\section{Representation comparison of SEnet and SEnetv2}
Researchers have asserted that utilizing aggregated modules containing multiple convolution operations through branching inputs can be more effective than opting for deeper networks or wider layers \cite{resnet,wrns,resnext}. These models becomes adept at capturing intricate spatial representations, particularly within the convolutional layers of the aggregated modules.

\begin{figure}[hbt!]
    \centering
    \includegraphics[width=0.48\linewidth]{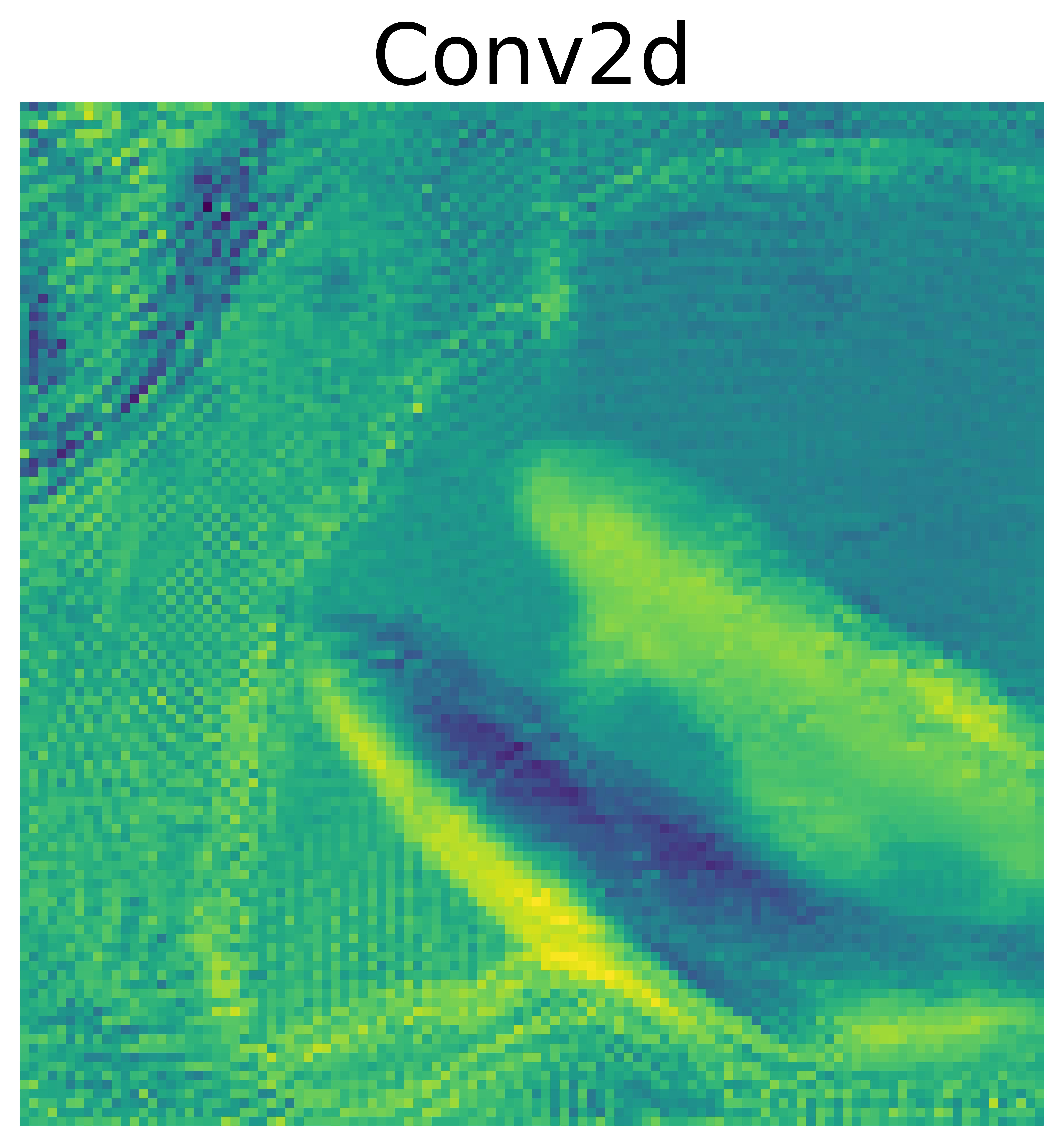}\hfill
    \includegraphics[width=.48\linewidth]{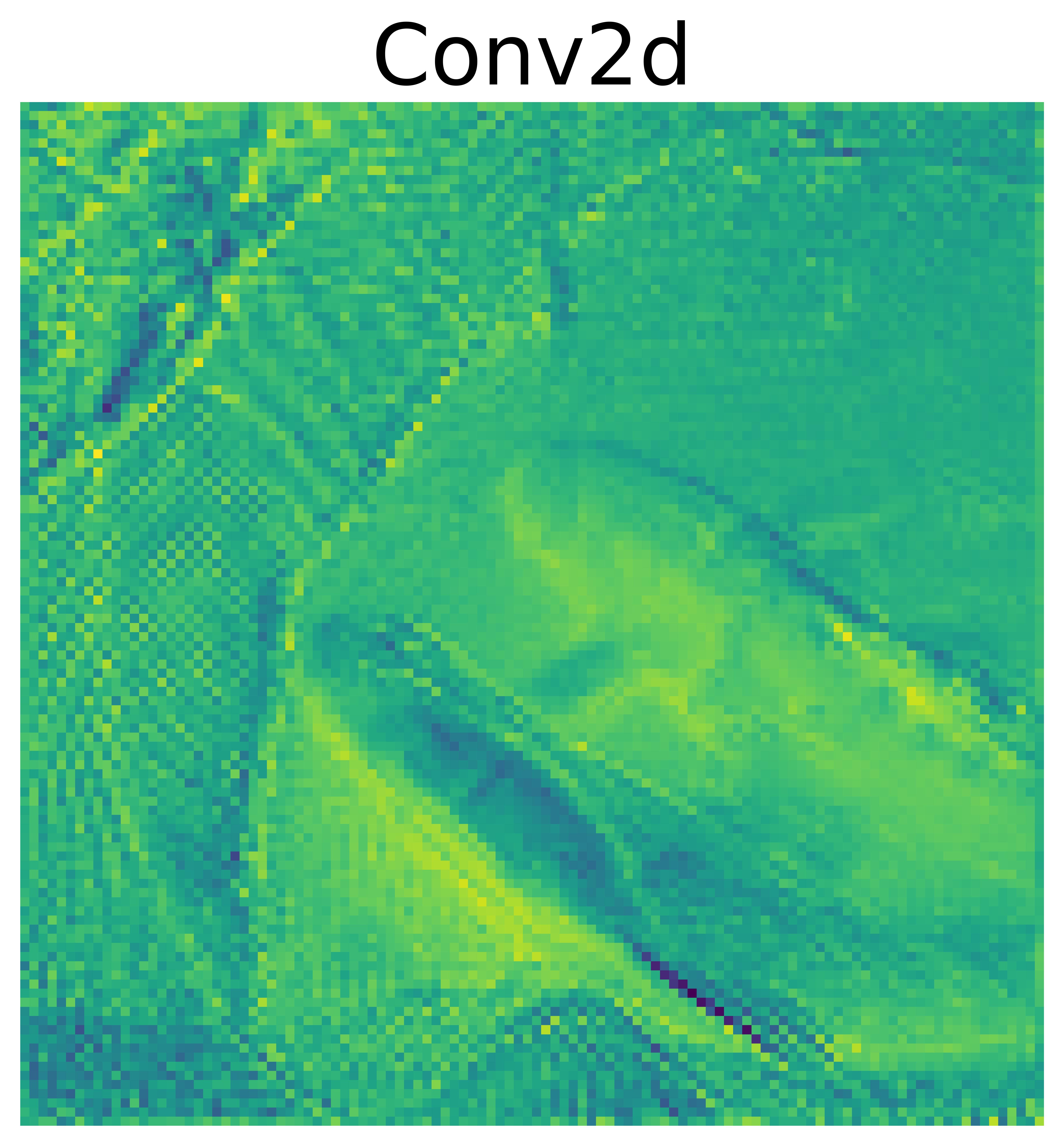}
    \caption{This explains the learned activation kernels for particular input dataset. We compare the learned values for the first convolution of SE resnet and the proposed SE resnet V2 models.}
    \label{fig:explainable}
\end{figure}

The images showcased in Figure \ref{fig:explainable} illustrate the acquired representations for the initial convolutional layer of both the SEnet and the proposed SEnetV2 architectures. These learned representations have been captured following a comprehensive training process of 50 epochs. The training initiates with an initial learning rate of 0.01, and every 15 epochs, this value is decayed by a factor of 0.1. This systematic approach enables the model to optimize its learning process effectively.

Upon close examination of the images, a notable distinction emerges which can be attributed to the fact that the consolidated information within the newer SEnet architecture is transmitted across a more extensive array of layers, thus facilitating a richer and more diverse learning process in comparison to SEnet.

\section{Methodology}

As elaborated in the preceding sections, the proposed network engages in the acquisition of spatial representations through its convolutional layers, embraces channel-wise representations via its squeeze and excitation layer, and gains enhanced channel-wise global representations through its aggregated layers. This SENetV2 architecture bears resemblance to ResNeXt, as it integrates an aggregated structure within the residual module of SENet. 

To facilitate a clearer understanding of the module, we apply the proposed approach in conjunction with the widely acknowledged Resnet \cite{resnet}. The residual module has emerged as a foundational model for evaluating and emphasizing the significance of network depth. This module incorporates shortcut connections that enable the bypassing of one or more layers. These layers are subsequently concatenated with the shortcut connection. In the elemental version of the residual module, with input denoted as 'x', the operations responsible for modifying the input, including batch normalization and dropout, are symbolized by 'F()'. The mathematical expression representing the residual module is as follows

\begin{equation}
    Resnet = x+F(x)
\end{equation}

In the context of the aggregated residual module, the input is directly concatenated without undergoing any alterations, given the utilization of branched convolutions. The resulting mathematical formulation is presented as follows

\begin{equation}
    ResneXt = x + \sum_{}F(x)
\end{equation}

The foundational aspect of the proposed module draws inspiration from the Squeeze and Excitation Network. This network module amalgamates two distinct techniques: Squeeze and Excitation. The Squeeze module is characterized by the input being subjected to a squeezing process through the utilization of fully connected (FC) layers. The output from the convolutional layer is directed into a global average pooling layer to generate a channel-wise input. This input is subsequently fed into an FC layer, which employs a reduction size. Conversely, the Excitation component of the module involves the application of an FC layer without any reduction, restoring the input to its original form. This FC layer is succeeded by a scaling operation, wherein the output is subjected to channel-wise multiplication with the feature map. The final output is subsequently rescaled to align with its original shape. The formulation of the squeeze and excitation operation for the residual module can be expressed as

\begin{equation}
    SEnet = x + F(x \cdot Ex(Sq(x)))
\end{equation}

Within this context, the 'Sq' function denotes the Squeeze operation, encompassing an FC layer with a reduction size denoted as 'r'. Subsequent to the 'Sq' operation, the 'Ex' operation, signifying the excitation operation, is executed. This process aims to restore the channel-wise modified inputs to their initial shape, devoid of any reduction. Following this, a scaling operation is enacted, employing the input to restore it to its original form. This scaled output is then concatenated with the input in the residual module.

To optimize the performance and maintain the fundamental premise of squeezed excitation, we scrutinized the impact of various group sizes in the squeezed format. Our experimentation led us to opt for a cardinality value of 4 unlike resnext which has 32. This cardinality is conducive to engaging the core essential features with the excitation layer, without resulting in unnecessary enlargement of the module leading to increase of model parameters. This strategic choice allows the module to effectively learn global representations while preserving an efficient structure.

In the squeeze operation, the FC layer with a reduced size operates on the output generated from global average pooling. This transformation facilitates the preservation of essential features, enabling them to traverse through the module and thereby amplifying the network's representational capabilities. In line with this, we put forth the concept of enhancing the cardinality of the FC layer within this operation. By introducing a new type of layers, multibranch dense layers alongside reduction, the model is equipped to learn a broader range of global representations throughout the network.

\begin{figure}
    \centering
    \includegraphics[width=0.9\linewidth,height=6cm]{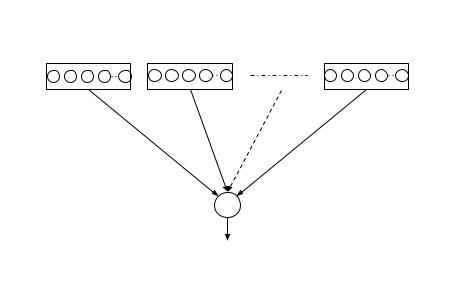}
    \caption{Figure represents the proposed aggregated fully connected layers within SEnetV2. This combines with concatenation at the end. The compressed conv layers after squeeze is fed as input to this module. The layers consists of reduction size 32 with the cardinality 4.}
    \label{fig:aggregated}
\end{figure}

To illustrate, the aggregated layers within the squeeze operation are concatenated and subsequently conveyed to the FC layer, as illustrated in Figure \ref{fig:aggregated}. Following this, the output from the FC layer is subjected to multiplication with the input layer of the module, resulting in dimension restoration. This final output undergoes a scaling operation akin to SENet. This sequence of operations within a residual module can be depicted as follows

\begin{equation}
    SEnetV2 = x + F(x \cdot Ex(\sum_{}Sq(x)))
\end{equation}

\begin{table*}[hbt!]
    \centering
    \setlength{\tabcolsep}{10pt}
    \renewcommand{\arraystretch}{1.5}
    \begin{tabular}{c | c | c}
        \hline
        Resnet-50 & SEnetV2-50 & SEneXtV2-50 \\[3pt]
        \hline
        \multicolumn{3}{c}{conv, 7x7, 64, stride 2} \\
        \hline
        \multicolumn{3}{c}{max pool, 3x3, stride 2} \\
        \hline
        $\begin{bmatrix}
             conv,1x1,64 \\[3pt]
             conv,3x3,64 \\[3pt]
             conv,1x1,256 \\[3pt]
            \end{bmatrix} \times 3 $ & 
        
        $\begin{bmatrix}
             conv,1x1,64 \\[3pt]
             conv,3x3,64 \\[3pt]
             conv,1x1,256 \\[3pt]
             fc,[8,256] \times 4 \\[3pt]
        \end{bmatrix}  \times 3 $ &
        
        $\begin{bmatrix}
             conv,1x1,128 \\[3pt]
             conv,3x3,128 \ C=32 \\[3pt]
             conv,1x1,256 \\[3pt]
             fc,[8,256] \times 4 \\[3pt]
        \end{bmatrix}  \times 3 $\\
        \hline
        $\begin{bmatrix}
             conv,1x1,128 \\[3pt]
             conv,3x3,128 \\[3pt]
             conv,1x1,512 \\[3pt]
            \end{bmatrix} \times 4 $ & 
        
        $\begin{bmatrix}
             conv,1x1,128 \\[3pt]
             conv,3x3,128 \\[3pt]
             conv,1x1,512 \\[3pt]
             fc,[16,512] \times 4 \\[3pt]
        \end{bmatrix}  \times 4 $ &
        
        $\begin{bmatrix}
             conv,1x1,256 \\[3pt]
             conv,3x3,256 \ C=32 \\[3pt]
             conv,1x1,512 \\[3pt]
             fc,[16,512] \times 4 \\[3pt]
        \end{bmatrix}  \times 4 $\\
        \hline
        $\begin{bmatrix}
             conv,1x1,256 \\[3pt]
             conv,3x3,256 \\[3pt]
             conv,1x1,1024 \\[3pt]
            \end{bmatrix} \times 6 $ & 
        
        $\begin{bmatrix}
             conv,1x1,256 \\[3pt]
             conv,3x3,256 \\[3pt]
             conv,1x1,1024 \\[3pt]
             fc,[32,1024] \times 4 \\[3pt]
        \end{bmatrix}  \times 6 $ &
        
        $\begin{bmatrix}
             conv,1x1,512 \\[3pt]
             conv,3x3,512 \ C=32 \\[3pt]
             conv,1x1,1024 \\[3pt]
             fc,[32,1024] \times 4 \\[3pt]
        \end{bmatrix}  \times 6 $\\
        \hline
        $\begin{bmatrix}
             conv,1x1,512 \\[3pt]
             conv,3x3,512 \\[3pt]
             conv,1x1,2048 \\[3pt]
            \end{bmatrix} \times 3 $ & 
        
        $\begin{bmatrix}
             conv,1x1,512 \\[3pt]
             conv,3x3,512 \\[3pt]
             conv,1x1,2048 \\[3pt]
             fc,[64,2048] \times 4 \\[3pt]
        \end{bmatrix}  \times 3 $ &
        
        $\begin{bmatrix}
             conv,1x1,1024 \\[3pt]
             conv,3x3,1024 \ C=32 \\[3pt]
             conv,1x1,2048 \\[3pt]
             fc,[64,2048] \times 4 \\[3pt]
        \end{bmatrix}  \times 3 $\\
        \hline
        \multicolumn{3}{c}{global average pool, 1000-d fc, softmax} \\
        \hline
    \end{tabular}
    \caption{The shapes and operations along with groups (C) and aggregated FC layers depicted with the cardinality values. (\textbf{left}): Resnet-50, (\textbf{Middle}): SENetV2-50 and (\textbf{Right}): SENeXtV2-50. The fc indicates the output dimension of the two fully connected layers in SENetV2.}
    \label{tab:se_design}
\end{table*}

\section{Squeeze aggregated excitation resnet}

Since the proposed module is built upon the foundation of the existing SENet, it inherits the same fundamental characteristics. This module is designed to seamlessly integrate into various network architectures. Specifically, the output layer of standard architectures can be effortlessly directed to the proposed module. In our implementation, we adopt the residual network as the host architecture. This choice is rooted in the widespread use of residual networks as a benchmark for testing various models \cite{resnext,biglittlenet,senet}, which simplifies the evaluation process when experimenting with existing architectures.

\begin{figure}
    \centering
    \includegraphics[width=0.95\linewidth,height=8cm]{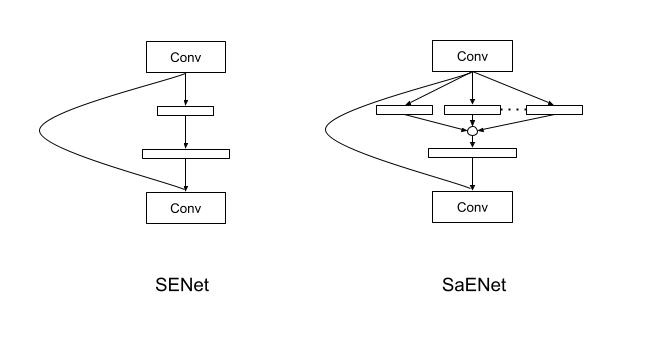}
    \caption{The figure illustrates a comprehensive comparison between the SENet module and the proposed SENetV2 module. In the figure, the smaller rectangles denote fully connected (FC) layers or dense layers, while the convolutional layers are represented as 'Conv'. Both the squeeze operation and the subsequent excite or aggregated excite operation are always combined with the input, ensuring the restoration of the original shape. Multibranch dense layers are concatenated in the circle.}
    \label{fig:sevssae}
\end{figure}

The detailed explanation of the proposed SaE (Squeeze Aggregated Excitation) module primarily focuses on its implementation within the ResNet architecture. Residual modules, which include a shortcut connection appended after specific layers, serve to convey the learned information of the network. This mechanism facilitates the construction of deeper networks while mitigating issues associated with the vanishing gradient problem. By integrating the SaE module into the residual module without altering the architecture, we take advantage of the inherent information-passing capacity of the residual module.

The selection of the residual module as the host is grounded in its consistent information propagation, which aids in enhancing the network's learning process \cite{resnext,senet}.

The squeeze operation employed within the SaE module involves the summation of output from all branched FC layers, followed by the excitation operation. A comparison between the SE (Squeeze and Excitation) module and the proposed SaE module within the context of the ResNet module is depicted in Figure \ref{fig:sevssae}. This illustration provides an insight into the intricate functioning of the Squeeze and Excitation operation.

The squeeze operation is initiated after the global average pooling layer, which extracts channel-wise statistics. This channel-wise information is subsequently funneled into the squeeze operation, where the input is dimensionally reduced. This is succeeded by the excitation layer. Within the architecture, the residual module is constructed by repeating a convolutional layer after specific intervals, forming a structured module. These modular units are repeated periodically to ensure the learned gradients are effectively propagated without succumbing to the vanishing gradient problem that can occur with deeper networks. The residual network is fundamentally rooted in the architecture of the VGG network \cite{vgg2014} and the ResNet \cite{resnet}, establishing a robust base for its design and functionality.

\section{Implementation}
We evaluated the effectiveness of the proposed method by conducting comparisons with state-of-the-art architectures. To facilitate this comparison, we employed multiple datasets: CIFAR10, CIFAR100 and ImageNet. In our experiments, we focused on implementing the proposed module within the ResNet architecture. The comparison between the residual network incorporating the proposed Squeeze Aggregated Excitation (SaE) module and other architectures, including ResNet and ResNeXt, is presented in Table \ref{tab:se_design}.

Both the proposed SENetV2 and the Aggregated Residual Network share the characteristic of the SaE module while also incorporating their respective architectural features. To maintain stable training dynamics and improve convergence, all convolutional layers in both architectures are coupled with batch normalization. Among various normalization techniques, batch normalization stands out due to its effectiveness in addressing the issue of internal covariate shift.

\section{Experiments}

The proposed method was empirically evaluated on two distinct datasets: CIFAR-10, CIFAR-100, and a customized version of Imagenet. In the experiments involving CIFAR-10 and CIFAR-100 datasets, we employed the resnet-18 and resnet-50 models for testing, respectively. For the CIFAR-10 dataset, input images with dimensions of 70x70 were utilized, and the model was trained for a total of 50 epochs. On the other hand, for CIFAR-100, input images with dimensions of 140x140 were used with the resnet-50 model, and the training process spanned 40 epochs. We use the Adam optimizer and categorical cross-entropy loss function, with a batch size of 16. The experiments were conducted using NVIDIA P100 GPUs for CIFAR-10 and CIFAR-100 datasets, and NVIDIA H100 GPUs for the computations involving the Imagenet dataset.

In the case of the Imagenet dataset, specific modifications were applied, which will be detailed in the subsequent sections. For training on the Imagenet dataset, input images of size 224x224 were organized into batches of size 256. The Stochastic Gradient Descent (SGD) optimizer was employed, incorporating a momentum of 0.9 and a weight decay of 1e-4.

\subsection{CIFAR-10}
Certainly, the CIFAR-10 dataset is a widely used benchmark dataset in the field of computer vision. It consists of 50,000 training and 10,000 testing images of size 32x32 in 10 different classes. The results are presented in a table \ref{tab:cifar10res}.

\begin{table}[hbt]
    \centering
    \begin{tabular}{c|c|c}
    \hline
    Models & Model Parameters & Accuracy (Top1)\\
    \hline
    Resnet & 8.06M & 77.38\\
    SE Resnet & 8.41M & 77.79\\
    \textbf{SE ResnetV2} & 9.46M & \textit{78.60}\\ 
    MobileNet & 3.23M & \textbf{81.35}\\
    \hline
    \end{tabular}
    \caption{Results obtained from experimenting on various models using CIFAR-10 dataset.}
    \label{tab:cifar10res}
\end{table}
The results clearly indicate that it enhances the performance of models with a similar structure. While it might lag behind MobileNet, it has the potential to boost accuracy with only a marginal increase in model parameters.

\subsection{CIFAR-100}
The experimentation was conducted on the CIFAR-100 dataset, which encompasses 100 distinct classes, with each class comprising 6000 images. Among these images 50,000 for training and 10,000 images used for testing. Our methodology was tested on this dataset, and the ensuing results have been organized and presented in a tabular format.

\begin{table}[hbt]
    \centering
    \begin{tabular}{c|c|c}
    \hline
    Models & Model Parameters & Accuracy (Top1)\\
    \hline
    Resnet & 23.62M & 61.72\\
    SE Resnet & 24.90M & 62.26\\
    \textbf{SE ResnetV2} & 28.67M & \textit{63.52}\\ 
    MobileNet & 3.24M & \textbf{65.40}\\
    \hline
    \end{tabular}
    \caption{Results obtained from experimenting on various models using CIFAR-100 dataset.}
    \label{tab:cifarres}
\end{table}

As indicated in Table \ref{tab:cifarres}, a noteworthy observation can be made regarding the performance of SEnet and the proposed SEnetV2 in comparison to the vanilla ResNet. The results reveal that in specific instances, such as the SENetV2, the proposed network exhibits superior performance. Notably, despite a difference of 5 million model parameters compared to the vanilla ResNet, having a similar depth is sufficient instead of opting for a deeper network.

\subsection{Imagenet results}
For the experiments conducted on the custom modified ImageNet dataset, a strategic approach was taken due to computational limitations. Recognizing the constraints, the dataset was customized to contain a reduced number of images. This is similar to Caltech 256 dataset where average images per class is 119 with the least of 80 in a class. For the purposes of the experiment, a subset of 250 images from Imagenet train was selected to represent the modified ImageNet dataset. The same validation set was employed for evaluating the trained model on this modified dataset, while the previously mentioned data transformations were applied as well.

\begin{table}[hbt]
    \centering
    \begin{tabular}{c|c|c}
        \hline
        Models & Top-1 & Top-5 \\
        \hline
        Resnet & 21.8740 & 43.4680 \\
        SEnet \cite{senet} & 22.1860 & 43.9180 \\
        SEnetV2 & \textbf{22.2820} & \textbf{43.9680} \\
        \hline
        ResneXt & 22.8000 & 44.7600 \\
        SEneXt \cite{senet} & 24.6640 & 47.7960 \\
        SEneXtV2 & \textbf{24.9200} & \textbf{47.9220} \\
        \hline
    \end{tabular}
    \caption{Results for the various models tested on modified Imagenet dataset. Train dataset consists of 250 images in the imagenet with full validation dataset (50 images per class). }
    \label{tab:imagenetres}
\end{table}

The experimental outcomes for the models are organized and presented in Table \ref{tab:imagenetres}. Upon analyzing the results, it is evident that the proposed module yields promising outcomes in terms of top-1 and top-5 accuracy in comparison to vanilla resnet and SE resnet. Notably, the SENetV2 achieves commendable top-1 and top-5 accuracy by classifying better compared to other experiments. 

\section{Conclusion}

In this paper, we present the SaE module, an advanced iteration of the SENet aimed at augmenting the model's representational capacities. Our thorough experimentation substantiates the superior performance of SENetV2 over existing models. Moreover, we have introduced multi-branch fully connected layer to enhance global representation learning. We are confident that the amalgamation of spatial, channel-wise, and global representations within the network's structure can substantially contribute to refining feature acquisition. Ultimately, our proposed network holds the potential to elevate feature extraction in diverse vision-centric domains.

{\small
\bibliographystyle{ieee_fullname}
\bibliography{PaperForReview}
}

\end{document}